\documentclass[10pt,twocolumn,letterpaper]{article}

\usepackage{iccv}
\usepackage{times}
\usepackage{epsfig}
\usepackage{graphicx}
\usepackage{amsmath}
\usepackage{amssymb}
\usepackage{multirow}
\usepackage{ulem}
\usepackage{arydshln}
\usepackage{bm}
\usepackage{booktabs}
\usepackage[table]{xcolor}
\usepackage{makecell}
\usepackage{ragged2e}
\usepackage[font=small,labelfont=bf]{caption}

\usepackage[pagebackref=true,breaklinks=true,letterpaper=true,colorlinks,bookmarks=false]{hyperref}

\iccvfinalcopy 

\def\iccvPaperID{3321} 

\ificcvfinal\fi

\let\oldtwocolumn\twocolumn
\renewcommand\twocolumn[1][]{%
   \oldtwocolumn[{#1}{
    \setlength{\abovecaptionskip}{0.cm}
    \vspace{-0.1in}
    \begin{center}
   \includegraphics[width=1\textwidth]{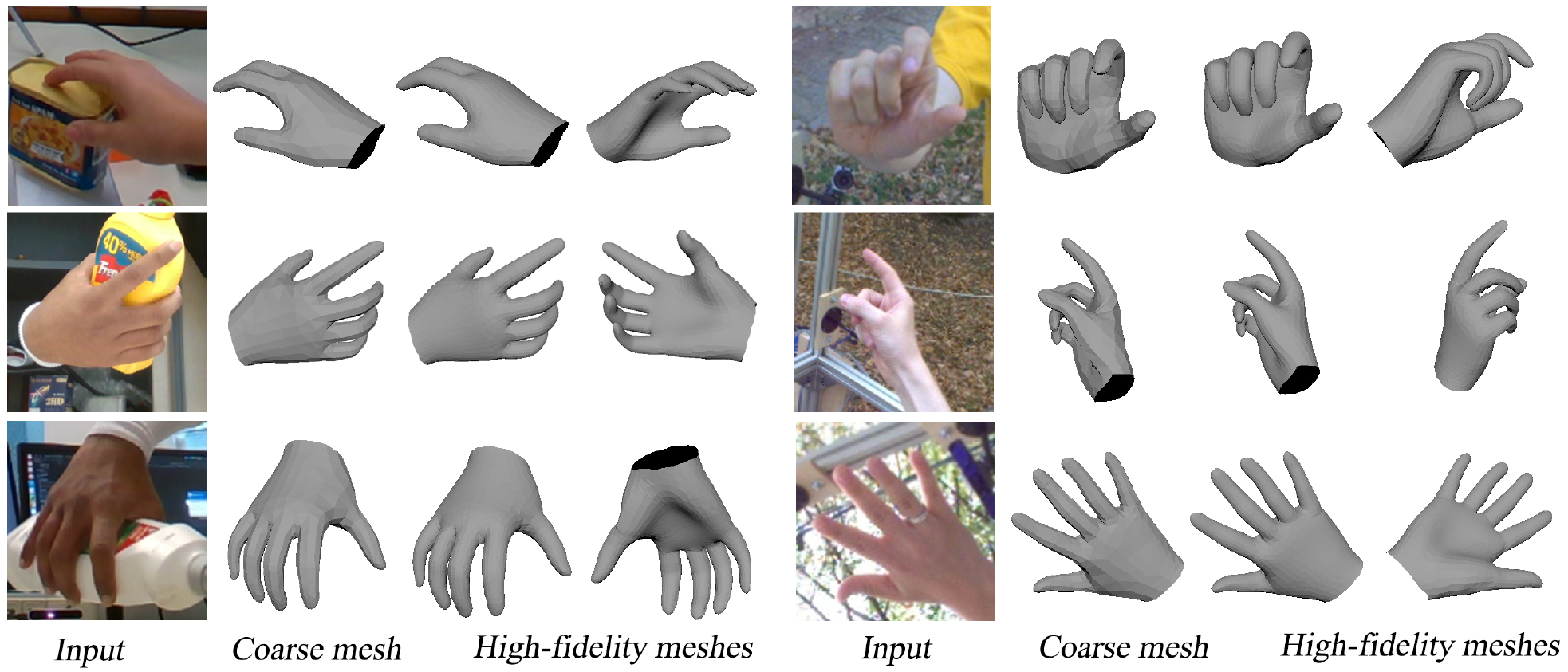}
    \captionof{figure}{We propose a novel Image-to-UV prediction network (I2UV-HandNet) that estimates accurate and high-fidelity hand mesh from a single RGB image. Here, we present example results on the HO-3D dataset (left) and the FreiHAND dataset (right). From left to right: input, estimated coarse mesh, estimated high-fidelity mesh (in two viewpoints).}
\label{fig:pipeline}
\end{center}
}]
}

\begin{document}

\vspace{-0.1in}
\title{I2UV-HandNet: Image-to-UV Prediction Network for Accurate and High-fidelity 3D Hand Mesh Modeling}

\author{
Ping Chen$^{1}$\quad Yujin Chen$^{2,3}$\quad Dong Yang$^{1}$\quad  Fangyin Wu$^{1}$\quad  Qin Li$^{1}$\quad  Qingpei Xia$^{1}$\quad  Yong Tan$^{1}$\\ 
	$^{1}$IQIYI Inc.\quad
	$^{2}$Wuhan University\quad
	$^{3}$Technical University of Munich\\
	{\tt\small \{redcping, terencecyj\}@gmail.com\quad \{yangdong01, wufangying, liqin01, xiaqingpei, tanyong\}@qiyi.com}\\
}

\maketitle

\begin{abstract}
Reconstructing a high-precision and high-fidelity 3D human hand from a color image plays a central role in replicating a realistic virtual hand in human-computer interaction and virtual reality applications. 
The results of current methods are lacking in accuracy and fidelity due to various hand poses and severe occlusions.
In this study, we propose an I2UV-HandNet model for accurate hand pose and shape estimation as well as 3D hand super-resolution reconstruction.
Specifically, we present the first UV-based 3D hand shape representation. To recover a 3D hand mesh from an RGB image, we design an AffineNet to predict a UV position map from the input in an image-to-image translation fashion.
To obtain a higher fidelity shape, we exploit an additional SRNet to transform the low-resolution UV map outputted by AffineNet into a high-resolution one. 
For the first time, we demonstrate the characterization capability of the UV-based hand shape representation.
Our experiments show that the proposed method achieves state-of-the-art performance on several challenging benchmarks.
\vspace{-0.2in}
\end{abstract}

\section{Introduction}
Observing and understanding the human hand has been an important task in computer vision and human-computer interaction, with applications from gesture recognition to augmented reality (AR) and virtual reality (VR).  
Recently, we have witnessed significant progress in 3D hand pose and shape estimation \cite{boukhayma20193d,cai2018weakly,chen2019so,Ge_2019_CVPR,simon2017hand,zimmermann2017learning,zhou2017towards}, driven by efforts in large-scale data collection and annotation \cite{zhang20163d,zimmermann2017learning,zimmermann2019freihand}, coupled with the development of 3D representations and learning methods \cite{bronstein2017geometric,qi2017pointnet}.
This has led to remarkable advances in 3D hand understanding from a single-view color image.

Due to the lack of hand surface data, most of the earlier works study 3D pose estimation by estimating 3D joint location from a single image \cite{athitsos2003estimating,cai2018weakly,iqbal2018hand,spurr2020weakly,zimmermann2017learning}.
However, the sparse joints representation cannot meet the needs of many applications such as interacting a virtual hand with an object in some immersive VR scenarios \cite{holl2018efficient}. 
To better display the hand surface, previous approaches regress a parametric hand model (MANO) \cite{romero2017embodied}  with articulated and nonrigid deformations \cite{boukhayma20193d,hampali2020honnotate,kolotouros2019convolutional,mueller2019real,Zhang_2019_ICCV}.
Although it is easy to use CNN to predict the MANO parameters from RGB input and use 3D annotations to supervise this regression process \cite{hasson2019learning,zimmermann2019freihand}, this high-dimensional nonlinear regression limits the accuracy of reconstructed hands.
Then regression-based methods introduce various intermediate representations to guide the training process. In these methods, the 3D hand reconstruction is decomposed into two stages, that first regresses a set of intermediate representations such as 2D keypoints, masks, or 3D keypoints, then predicts the model parameters from these intermediate representations \cite{Zhang_2019_ICCV,zhou2020monocular}.
The performance of these works largely depends on the design of these intermediate representations as well as the usage of reasonable supervision terms.
More recently, \cite{Ge_2019_CVPR,Kulon_2020_CVPR} remove the dependence of parametric model prior and directly regress the 3D coordinates of mesh vertices. 
Even though good performances are shown, the above methods, which estimate model parameters or vertices coordinate from high-dimensional encoded features, break the spatial relationship contained in the original pixel space. 
\cite{moon2020i2l} proposes to predict a 1D heatmap for each mesh vertex coordinate and achieves state-of-the-art performance, but it only preserves spatial information in feature transformation while its vertex-wise output is still a discrete 3D representation.
Different to above 3D representation and learning method, we propose to use UV position map \cite{feng2018joint} as the hand representation in this work.

Inspired by recent 3D body recovery methods that map a 3D mesh of the human body into a UV map representation \cite{alldieck2019tex2shape,yao2019densebody}, we propose to represent 3D hand surface in UV space and train a neural network to predict 3D hand shape from a single RGB input. 
The usage of UV representation enables an efficient network to directly regress the hand surface, without relying on any model prior or intermediate representations. 
To properly predict the UV position map from the RGB input, we present AffineNet that addresses the single-view 3D reconstruction issue in an image-to-image translation task.
Traditional image-to-image conversion pipelines are designed for tasks (such as appearance conversion or semantic segmentation) with good spatial alignment between the input and the output \cite{wang2018high,zhu2017unpaired}.
However, in our setup, the hand shape displayed by the UV position map is different from that in the input RGB image. To address this problem, we propose a novel affine connection module to align the encoded feature maps with the UV maps and then connect the aligned feature maps with the decoded feature maps. 
In AffineNet, hierarchical UV position maps and multi-level feature maps are employed, and multiple UV maps can be supervised at the training stage. For 3D pose estimation, we obtain a set of 3D keypoints from the output hand mesh via a pre-defined mapping.

Another advantage of the UV-based representation is that the dense UV position map enables reconstructing a 3D surface with more vertices by sampling in the valid area of the UV position map.
Motivated by this observation, we present a UV-based 3D hand super-resolution reconstruction module named SRNet to realize high-fidelity 3D hand reconstruction from the coarse 3D hand shape.
In order to make the best of the proposed hand UV position map representation, we restore high-fidelity hand shape by using a CNN to map the low-resolution UV position map into a high-resolution one. 
However, there lacks of high-fidelity hand surface data to supervise the learning of SRNet.
Thus, we construct a scan dataset called SuperHandScan to learn the SRNet.
We transfer the high-quality 3D hand scan and the registered coarse MANO model to high/low-resolution UV position maps, and then use those UV maps to train the SRNet.
Since the input of SRNet is a coarse hand mesh in UV-based representation, there is wide application scope for the SRNet, in other words, a well-trained SRNet can be used for mesh super-resolution reconstruction of any coarse hand mesh. 

In summary, we present an I2UV-HandNet model which consists of an AffineNet for 3D hand pose and shape estimation and an SRNet for hand mesh super-resolution reconstruction. 
Overall, the main contributions of this paper are summarized as follows:
\vspace{-0.2cm}
\begin{itemize}
\setlength{\itemsep}{0pt}
\setlength{\parsep}{0pt}
\setlength{\parskip}{0pt}
\setlength{\topsep}{0pt}
\setlength{\partopsep}{0pt}
\item [$\bullet$] To our best knowledge, we are the first to introduce UV map representation in 3D hand pose and shape estimation. Based on our novel representation, we propose an end-to-end network named AffineNet to predict hand mesh from a single color image.
\item [$\bullet$] For the first time in hand reconstruction, we propose SRNet, a hand mesh super-resolution reconstruction network to predict a high-fidelity hand mesh from a coarse hand mesh. 
\item [$\bullet$] Our method can predict accurate and high-fidelity hand meshes from RGB inputs. Experimental results show that our method surpasses other state-of-the-art methods on multiple challenging datasets.
\vspace{-0.4cm}
\end{itemize}

\begin{figure*}[t]
\vspace{-0.2in}
\begin{center}
\includegraphics[width=1\linewidth]{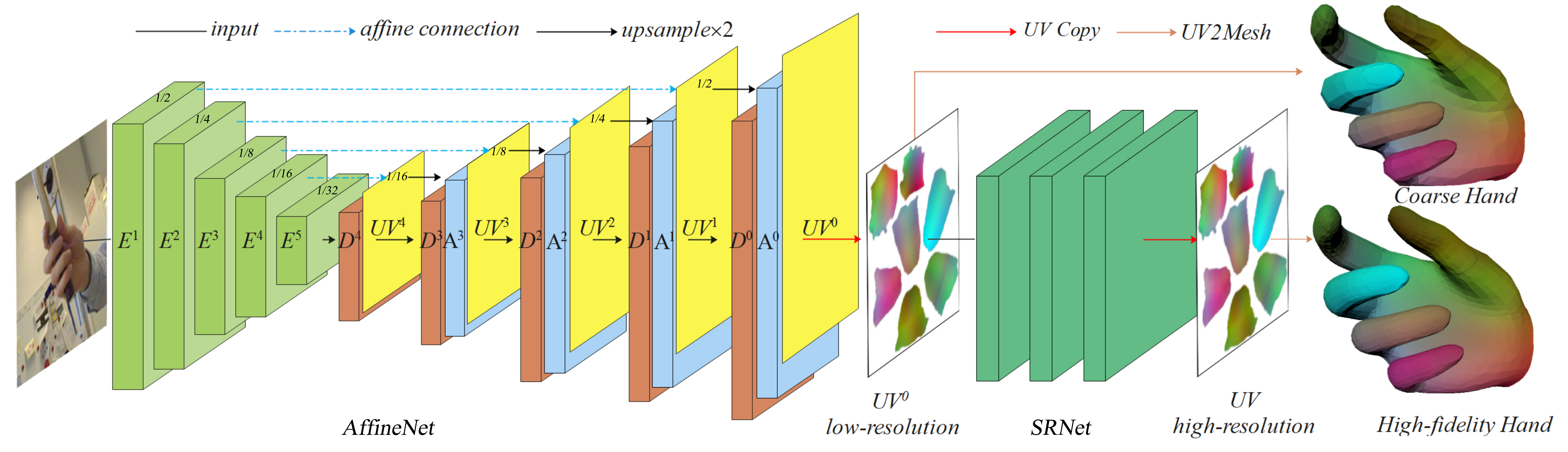}
\captionof{figure}{
Overview of the propose framework. The proposed I2UV-HandNet model, which consists of the AffineNet and the SRNet, enables to predict high-precision and high-fidelity hand meshes from a single-view color image. 
}
\label{fig:overall}
\end{center}
\vspace{-0.7cm}
\end{figure*}
\section{Related Work}
This section introduces related work on 3D hand pose estimation, hand pose and shape recovery, and dense shape representation.
Below, we compare our contribution with prior works.

\subsection{3D Hand Pose Estimation}
The task of 3D hand pose estimation aims to predict the 3D position of hand joints. Recently, estimating 3D hand pose from depth image or RGB image is well-explored. For works on depth-based hand pose estimation, please refer to \cite{armagan2020measuring,yuan2018depth}. Here, we mainly compare RGB-based hand pose estimation methods. 

Because the 3D joint annotations are hard to directly obtain from the 2D images, many methods make use of the correspondence of 3D joints and their 2D projections to boost 3D pose estimation.
\cite{simon2017hand} proposes to detect 2D hand keypoints of multi-view color images, and then these 2D detections are triangulated in 3D using multi-view geometry.
After the emergence of many datasets with 3D annotations \cite{zhang20163d,zimmermann2017learning}, methods are explored to directly regress 3D joints from RGB images by using 3D annotations to supervise network training \cite{zimmermann2017learning}.
Then, many methods follow this strategy and improve the performance either by introducing intermediate representations \cite{iqbal2018hand} or using more supervision terms \cite{cai2018weakly, spurr2020weakly}. 
Recently, a 3D joint is represented as three 1D heatmaps in \cite{moon2020i2l}, where the results are also convincing. In this paper, we only use the output 3D joints to help evaluate the performance of hand mesh modeling.


\subsection{Hand Pose and Shape Estimation}
Since sparse keypoints have a limited representation of 3D shapes, recent works combine sparse pose estimation with dense shape reconstruction to provide a more comprehensive shape representation. Methods in this area can be split into two categories with differences in the shape prior model used or not.

To solve the problem that 2D image lacks sufficient depth information and shape knowledge for 3D shape recovery, parametric shape models (e.g., 3DMM for face \cite{blanz1999morphable}, SMPL for body \cite{loper2015smpl}, and MANO for hand \cite{romero2017embodied}), which are built using 3D scan data, use low-dimension parameters to represent the complex 3D surface.
Recent works \cite{boukhayma20193d,hampali2020honnotate,kolotouros2019learning,wang2020rgb2hands,Zhang_2019_ICCV} integrate a parametric-based hand model (MANO) with the end-to-end deep network for hand pose and shape estimation.
The basic idea is to regress MANO parameters from the input image and then recover 3D joints and shape according to the regressed parameters, where 3D joints and meshes or fitted MANO parameters are used to supervise network training \cite{boukhayma20193d,hasson2019learning,Zhang_2019_ICCV, zhou2020monocular}. 
There are also method which attempt to remove 3D supervision \cite{chen2021self,spurr2020weakly} or recover 3D shape from 3D pose \cite{choi2020pose2mesh}.

Although the parametric model brings 3D shape priors, estimating model parameters from an RGB image breaks the spatial relationship between 2D pixels. 
To address this issue, I2L-MeshNet \cite{moon2020i2l} predicts 1D heatmaps for each mesh vertex coordinate instead of directly regressing the parameters. 
\cite{Ge_2019_CVPR} and \cite{Kulon_2020_CVPR} regress per-vertex position via graph convolution networks (GCNs).
Unlike them representing 3D hands in 3D space, we represent the surface of 3D hands by 2D UV maps which can be mapped from the input image in an image-to-image translation fashion like \cite{isola2017image}.

For the number of vertice of the output mesh, \cite{Kulon_2020_CVPR, moon2020i2l} use 778 vertices and use the same mesh topology as MANO, and \cite{Ge_2019_CVPR} outputs a mesh with 1,280 vertices via GCN.
DeepHandMesh \cite{moon2020deephandmesh} regresses a high-fidelity mesh with 12,553 vertices, but it needs to be trained (or pre-trained) on the data captured from the controlled environment and still suffers from the highly non-linear regression problem due to it represents the hand via parameter.
In this work, we propose a more general solution for high-fidelity hand reconstruction, where we input a low-resolution UV position map (MANO-level hand mesh) and output a high-resolution UV position map (high-fidelity hand mesh).

\subsection{Dense Shape Representation}
Although representing 3D shape via parametric models or 3D triangular mesh is straightforward and easy to employ supervision, there are works \cite{alldieck2019tex2shape,feng2018joint,yao2019densebody,zeng20203d,zhang2020object} which propose to represent 3D surface in a denser fashion, i.e., UV representations are introduced to represent the image-to-surface correspondences and then powerful 2D CNN can be directly utilized to learn the image-to-UV mapping. 
The UV representation can be divided into IUV and UV location maps, where the same position on the IUV and the RGB image shows spatial consistency, while the UV location map is inconsistent with RGB. 
Here, we call this inconsistency a coordinate ambiguity.
The IUV representation is used in single-view 3D face reconstruction \cite{feng2018joint} and body reconstruction\cite{xu2019denserac,yao2019densebody}.
Recently, \cite{alldieck2019tex2shape,zeng20203d} combine IUV, UV position maps and SMPL model \cite{loper2015smpl} to reconstruct 3D human. 
Even though these methods achieve good results, the coordinate ambiguity of the RGB and UV position map is not well addressed. 
In this paper, we introduce UV position maps to the hand reconstruction task for the first time and propose to reduce the coordinate ambiguity via an affine connection module (Section~\ref{sub_sec:affinenet}).

\begin{figure}[t]
\vspace{-0.2in}
\begin{center}
\includegraphics[width=1\linewidth]{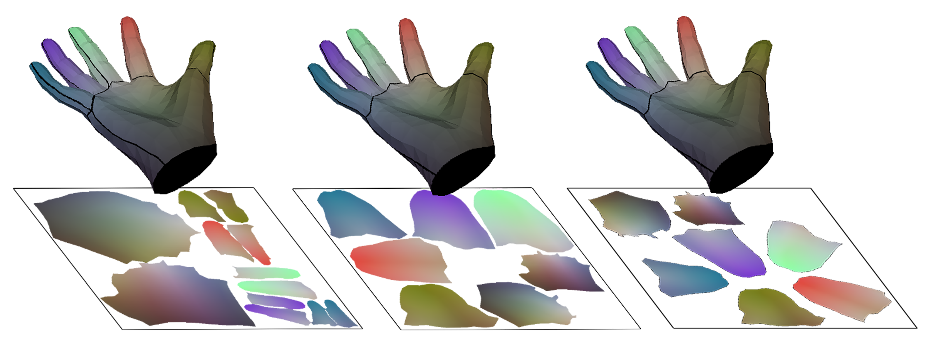}
\vspace{-0.2cm}
\captionof{figure}{Three UV unfolding forms in terms of different cutting and combination strategies.
}
\vspace{-0.3in}
\label{fig:2}
\end{center}
\end{figure}

\section{Method}
The proposed I2UV-HandNet model enables to represent 3D hand surface by UV position map (Section \ref{sec:representation}) and learn to estimate accurate and high-fidelity hand shape from a single-view image in a image-to-image translation fashion (Section \ref{sec:architecture} and \ref{sec:training_obj}).
In the following, we describe the proposed method in detail.

\subsection{3D Hand Representation}
\label{sec:representation}
\noindent \textbf{Parametric Hand Model.}\quad MANO is a parameterized hand model learned from hand scans. It defines a mapping from pose and shape parameters to a mesh of 778 vertices and 1538 faces, where the face topology is fixed to indicate the connection of the vertices in the hand surface. 
From the given mesh topology, a set of 21 joints can be directly formulated from the hand mesh. Here, we use the MANO model to infer 16 joints and obtain 5 fingertips according to pre-defined vertex indexes \cite{hasson2019learning}.\\
\textbf{Hand Surface as UV Position Map.}\quad Given a hand surface, such as the MANO hand mesh, we can unfold the surface into one UV map\footnote{\scriptsize \url{https://www.autodesk.com.sg/products/maya}}, which allows representing 3D surfaces as an image. Here, $U$ and $V$ denote the two axes of the image. 
The UV mapping defines the correspondence between the mesh vertices and the image pixels. 
Three UV mapping forms are shown in Figure~\ref{fig:2} in terms of different cutting and combination strategies, and the ablation comparison is conducted in Section~\ref{sub_sec:ablation}.
In our pipeline, the AffineNet directly outputs a UV position map from the input image, and the SRNet outputs a UV position map from a UV position map input, and then 3D hand meshes are recovered from UV position maps via the above-defined UV mapping.
To learn our model, the hand mesh annotations are transferred into UV maps to supervise the UV map prediction. 
Specifically, the hand mesh is spatially aligned with the corresponding RGB image using orthographic projection, so that the 3D hand mesh matches the 2D hand in the image plane. For each 3D vertex on the mesh, its 3D coordinate is mapped into the RGB channel value of a point in the UV position map \cite{yao2019densebody}. Interpolation is applied to generate continuous images. 
\subsection{I2UV-HandNet}
\label{sec:architecture}
As presented in Figure~\ref{fig:overall}, the proposed I2UV-HandNet model achieves accurate and high-fidelity hand shape estimation via an AffineNet to realize UV position maps prediction and an SRNet to restore high-resolution UV position maps.
\subsubsection{AffineNet}
\label{sub_sec:affinenet}
To predict the UV position map of the hand shape, an encoder-decoder mechanism is adopted to map the input image into a UV image. 
Similar to the U-net\cite{ronneberger2015u}, the AffineNet, consisting of a contracting path and an expansive path (as shown in Figure~\ref{fig:overall}).
Give a color image $I$ with a hand in its scope, a ResNet-50 backbone \cite{he2016deep} is used to encode the image into a series of encoded features $\{E^i|i=1,2,3,4,5\}$ with different resolutions. 
In the expansive path, each step upsamples the feature map and UV map prediction along with making use of the corresponding encoded feature maps, resulting in a series of decoded features $\{D^i|i=0,1,2,3,4\}$ and predicted UV position maps $\{I_{UV}^i|i=0,1,2,3,4\}$:
\begin{equation}
\left\{
\begin{array}{l}
    {D}^4 = f_{up}(E^5)\\
    {I}^4_{UV} = f_{con}(D^4)
\end{array}
\right.
\end{equation}
and
\begin{equation}
\left\{
\begin{array}{l}
    A^3 = f_{up}(f_{ac}(\pi(I^{4}_{UV}),E^{4}))\\
    {D}^3 = f_{up}(D^4),\\
    {I}^3_{UV} = f_{con}(A^3,D^3,f_{up}(I^{4}_{UV})),
\end{array}
\right.
\end{equation}
and
\begin{equation}
\left\{
\begin{array}{l}
    A^i = f_{up}(f_{ac}(\pi(I^{i+1}_{UV}),E^{i+1})),\\
    D^i=f_{up}(f_{con}(D^{i+1},A^{i+1},I^{i+1}_{UV})),\quad i=0,1,2\\
    {I}^i_{UV} = f_{con}(A^i,D^i,f_{up}(I^{i+1}_{UV})),
\end{array}
\right.
\end{equation}
Here, $E^i$ is the feature map encoded at the $i$-th pyramid level, ${A}^i$ is the UV-aligned feature by an affine transformation, $D^i$ is the feature map, $f_{up}$ indicates $2\times$ up-sampling, $f_{ac}$ indicates the affine connection operation, $f_{con}$ indicates convolutional layers, and $\pi$ indicates the projection from UV position map to image coordinate system. Smaller $i$ indicates a higher resolution.
We note that the affine connection $f_{ac}$ aligns encoding features to decoding features through an affine-operation before connecting them, where the affine-operation, similar to the STN \cite{jaderberg2015spatial}, is based on the 2D projection of each vertex coordinate in the currently predicted UV map.
We provide more details in the Appendix.
\subsubsection{SRNet}
Since the network of our 3D hand surface is represented by UV position maps, we propose to get a more refined hand surface via a super-resolution in UV image space. 
We propose an SRNet to transfer the low-resolution UV position map into a high-resolution map. The network architecture of the SRNet is similar to the super-resolution convolutional neural network (SRCNN) \cite{dong2015image}, but the input and output are UV position maps instead of RGB image. After regressing a high-resolution UV position map via the SRNet, a hand mesh with higher fidelity can be reconstructed (as shown in the right part of Figure~\ref{fig:overall}). More details about SRNet architecture are provided in the Appendix.

\subsection{Training Objective}
\label{sec:training_obj}
\subsubsection{Losses of the AffineNet}
\label{sec:loss_affine}
To learn the AffineNet, we enforce UV alignment $E_{UV}$, UV gradient alignment $E_{grad}$, and mesh alignment $E_{verts}$:
\begin{equation}
    E_{affine}=\lambda_1E_{UV}+\lambda_2E_{grad}+\lambda_3E_{verts}
\end{equation}
\textbf{UV Alignment.}\quad  We propose a UV alignment loss $E_{UV}$ base on the L1 distance between the ground truth UV position map $\hat{I}_{UV}$ and the output UV position map $I_{UV}$:
\begin{equation}
E_{UV}=\left|(I_{UV}-\hat{I}_{UV})\cdot M\right|
\end{equation}
Here, $M$ is UV map mask since only the valid region of the UV position map has corresponding region on the 3D hand surface.\\
\textbf{UV Gradient Alignment.}\quad The ideal hand surface should be continuous, so does the UV position maps. To this end, we introduce a UV gradient alignment to encourage the predicted UV position map share the same gradient with the ground-truth UV position map: 
\begin{equation}
\begin{aligned}
    E_{grad} = |\partial_u(I_{UV}\cdot M)-\partial_u(I^{\ast}_{UV}\cdot M)|\\
    +|\partial_v(I_{UV}\cdot M)-\partial_v(I^{\ast}_{UV}\cdot M)|
\end{aligned}
\end{equation}
where $\partial_u$ and $\partial_v$ are gradients along the U-axis and V-axis, respectively.\\
\textbf{Mesh Alignment.}\quad Apart from $E_{UV}$ and $E_{grad}$ that measure the shape reconstruction in 2D UV position map space, we also introduce a mesh alignment loss $E_{verts}$ to enforce the predicted 3D hand mesh to be closed with the ground truth one:
\begin{equation}
    E_{verts} = \frac{1}{N_{vert}}\sum_{i=1}^{N_{vert}}{|v_{i}-\hat{v}_{i}|}
\label{eq:vert}
\end{equation}
Here, $v_i$ and $\hat{v}_i$ are the 3D coordinate of the $i$-th vertex from the output mesh and the ground truth mesh, respectively. $N_{vert}$ indicates the number of vertices of the mesh. $\mathcal{L}_{L1}$ is the L1 loss function.

Since there are multiple predicted UV position maps, we employ $E_{affine}$ on multi-scale UV maps. When training the AffineNet, the last four UV maps are used with equal weights.

\subsubsection{Losses of the SRNet}
The output of the SRNet is UV position map is similar to the output of the AffineNet except that the SRNet can produce a UV map with higher resolution. Here, we adopt similar loss functions with AffineNet.
\begin{equation}
    E_{SR} = E_{UV\_SR} + E_{verts\_SR}
\end{equation}
Here, we replace the component in $E_{UV}$ with the corresponding component from the SRNet to formulate $E_{UV\_SR}$, e.g., the UV position map is replaced by the UV position map with higher resolution. Also, the $E_{verts\_SR}$ is formulated in the same manner.

\section{Experiments}
In this section, we first present datasets  (Section~\ref{sub:dataset}) and evaluation metrics (Section~\ref{sub:evaluation_metrics}), and implementation details (Section~\ref{sec:implementation}). Then, the overall performance of the proposed method and comprehensive analysis are presented (Section~\ref{sub:sota}, \ref{sub:sr_evla} and \ref{sub_sec:ablation}).  
\subsection{Datasets}
\label{sub:dataset}
\noindent \textbf{FreiHAND.}\quad The FreiHAND dataset \cite{zimmermann2019freihand} contains real-world hand data with various poses, object interactions, and varying lighting. It contains 130,240 training samples and 3,960 test samples.
Each training sample contains a single-view RGB image, annotations of MANO-based 3D hand joints and mesh, as well as camera pose parameters. 
The result of the test set is evaluated via an online submission system\footnote{\scriptsize \url{https://competitions.codalab.org/competitions/21238}}.
\\
\textbf{HO3D.}\quad The HO3D dataset \cite{hampali2020honnotate} is a recently released dataset that collects color images of a hand with object interactions. This dataset has 66,034 training samples, which consists of single-view RGB images, MANO-based hand joints and meshes, and camera poses. 
For the test set, 11,524 RGB images are provided along with the annotation of the detection bounding box. 
The objects in this dataset are mainly from the TCB-Video dataset \cite{choi2020pose2mesh}.
The results of the test set need to be evaluated through its online submission system\footnote{\scriptsize \url{https://competitions.codalab.org/competitions/22485}}.

\noindent \textbf{ObMan.}\quad The ObMan dataset \cite{hasson2019learning} is a large-scale synthetic dataset containing hand-object interaction images. It contains 141,550 training samples, 6463 validation samples, and 6285 test samples. Each sample has an RGB-D image, 3D hand joints, 3D hand mesh, object mesh as well as camera pose parameters.

\noindent \textbf{YT-3D.}\quad The YouTube-3D-Hands (YT-3D) dataset collects images of various real-world hand from YouTube and annotate those images via an automated collection system \cite{Kulon_2020_CVPR}.
The training set, which is generated from 102 selected videos, has 47,125 hand images with 3D joint and mesh annotation. The validation and test sets cover 7 videos and contain 1525 samples each.

\noindent \textbf{SHS.}\quad We build the SuperHandScan (SHS) dataset using a collection of high-quality 3D hand scans via a laser scanner.
The motivation of SHS is that the MANO hand mesh, which represents the hand surface via 778 vertices (with 1538 faces), can only show coarse surface information, but the UV-based method can produce a hand surface with more details. Thus, we obtain three times higher resolution hand meshes based on a collection of hand scans and the given MANO model.
Specifically, we first up-sample the original MANO hand mesh from 778 vertices (with 1538 faces) to 3093 vertices (with 6152 faces) using the edge-based unpooling method as \cite{wang2018pixel2mesh}.
Then, the iterative closest point (ICP) algorithm is used to register the upsampled 3D mesh to the 3D point cloud (from the scanner). 
Our SHS dataset provides 6000 scans with dense 3D point clouds and the corresponding hand meshes. The hand mesh, which has 3093 vertices and 6152 faces, is denser than MANO hand mesh, thus can supervise SRNet to learn higher quality hand mesh.

\noindent \textbf{HIC.}\quad The Hands in Action Dataset (HIC) \cite{tzionas2016capturing} contains images of hand-object interaction. 
Each sample has the RGB-D image, 3D object shape, and MANO-fitted hand shape. We use all of the samples to evaluate the SRNet.

\subsection{Evaluation Metrics}
\label{sub:evaluation_metrics}
To evaluate the performance of the proposed method, multiple metrics are used for hand pose estimation and mesh reconstruction.
The \textbf{Pose error} measures the average Euclidean distance between the predicted and the ground truth 3D joints.
The \textbf{Mesh error} measures the average Euclidean distance between the predicted and the ground truth mesh vertices.
The \textbf{Pose AUC} indicates the area under the curve (AUC) for the plot of the percentage of correct keypoints (PCK) and the \textbf{Mesh AUC} indicates the AUC for the plot of the percentage of correct vertices (PCV).
We also compare the \textbf{F-score} \cite{knapitsch2017tanks} which is the harmonic mean of the recall and precision between two meshes given a distance threshold. We report the F-score of mesh vertices at 5mm and 15mm by F@5mm and F@15mm. Following the recent works, we compare aligned prediction results with Procrustes alignment.

We use PSNR and RMSE to evaluate the performance of hand super-resolution reconstruction.
The \textbf{PSNR} indicates computes the peak signal-to-noise ratio, in decibels, between two images and is used as a quality measurement between the original and a reconstructed image.
The higher the PSNR, the better the quality of the reconstructed surface.
The \textbf{RMSE} (Root Mean Square Error in \textit{mm}) is the standard deviation of the residuals.
In this work, we use PSNR and RMSE to evaluate the difference between the rendered depth map and the corresponding ground truth.

\begin{table*}[]
\centering
\vspace{-0.1in}
\scalebox{0.8}{
\begin{tabular}{lcccccc}
\Xhline{1pt}
Method                   & Pose Error↓ & Pose AUC↑ & Mesh Error↓ & Mesh AUC↑ & F@5 mm↑ & F@15 mm↑ \\
\hline
Boukhayma \textit{et al.} \cite{boukhayma20193d} & 3.50         & 0.351     & 1.32        & 0.738     & 0.427   & 0.894    \\
Zimmermann \textit{et al.} \cite{zimmermann2019freihand} (Mean Shape)  & 1.71        & 0.662     & 1.64        & 0.674     & 0.336   & 0.837    \\
Zimmermann \textit{et al.} \cite{zimmermann2019freihand} (Mano Fit) & 1.37        & 0.730      & 1.37        & 0.729     & 0.439   & 0.892    \\
Hasson \textit{et al.} \cite{hasson2020leveraging}   & 1.33        & 0.737     & 1.33        & 0.736     & 0.429   & 0.907    \\
Spurr  \textit{et al.} \cite{spurr2020weakly} & 1.13 & 0.78 & - & - & - &-\\
Zimmermann \textit{et al.} \cite{zimmermann2019freihand} (MANO CNN)     & 1.10         & 0.783     & 1.09        & 0.783     & 0.516   & 0.934    \\
Kulon \textit{et al.} \cite{Kulon_2020_CVPR}  & 0.84        & 0.834     & 0.86        & 0.830      & 0.614   & 0.966    \\
Choi \textit{et al.} \cite{choi2020pose2mesh}  & 0.77        & -         & 0.78        & -         & 0.674   & 0.969    \\
Moon \textit{et al.} \cite{moon2020i2l} & 0.74        & 0.854     & 0.76        & 0.850     & 0.681   & \textbf{0.973}    \\
Ours (AffineNet) & \textbf{0.72} & \textbf{0.856} & \textbf{0.74} & \textbf{0.852} & \textbf{0.682} & \textbf{0.973}\\
\hline
Ours* (AffineNet) & 0.68 & 0.865 & 0.69 & 0.862 & 0.706 & 0.977\\
\Xhline{1pt}
\end{tabular}
}
\vspace{-0.1in}
\caption{Comparison of main results on the FreiHAND test set.
* indicates the system is trained on a combination of datasets.
}
\label{table:1}
\end{table*}

\begin{table*}[tb]
\vspace{-0.1cm}
\centering
\makebox[0pt][c]{\parbox{1.05\textwidth}{
\centering
\begin{minipage}[h]{0.67\textwidth}
\centering
\scalebox{0.74}{
\begin{tabular}{lcccccc}
\Xhline{1pt}
Method                 & Pose Error↓ & Pose AUC↑ & Mesh Error↓ & Mesh AUC↑ & F@5 mm↑ & F@15 mm↑ \\
\hline
Hasson \textit{et al.} \cite{hasson2020leveraging} & 1.14        & 0.773     & 1.14        & 0.773     & 0.428   & 0.932    \\
Hasson \textit{et al.} \cite{hasson2019learning}  & 1.10        & -         & -           & -         & 0.46    & 0.93     \\
Hampali \textit{et al.} \cite{hampali2020honnotate} & 1.07        & 0.788     & 1.06        & 0.790     & \textbf{0.506}   & 0.942    \\
Ours (AffineNet) & \textbf{0.99} & \textbf{0.804} & \textbf{1.01} & \textbf{0.799} & 0.500 & \textbf{0.943}\\
\hline
Ours$\dagger$ (AffineNet) & 1.04 & 0.793 & 1.09 & 0.782 & 0.484 & 0.935\\
Ours$*$ (AffineNet) & 0.81 & 0.838 & 0.84 & 0.831 & 0.577 & 0.970\\
\Xhline{1pt}
\end{tabular}
}
\vspace{-0.1in}
\caption{Comparison of main results on the HO3D test set.
$\dagger$ indicates cross-dataset evaluation.
$*$ indicates trained use extra data.
}
\label{table:ho3d}
\end{minipage}
\begin{minipage}[h]{0.33\textwidth}
\centering
\scalebox{0.8}{
\begin{tabular}{l|cc}
\Xhline{1pt}
Set              & RMSE↓ & PSNR↑ \\
\hline
Input            & 28.12 & 27.58 \\
Output (sampled) & 12.06 & 37.74 \\
Output           & \textbf{7.68}  & \textbf{39.18} \\
\Xhline{1pt}
\end{tabular}
}
\vspace{-0.1in}
\caption{Comparison of main results on HIC. Note that the depth maps are transferred into point clouds in world coordinates, and then the distance is computed between corresponding points.
}
\label{table:srnet}
\end{minipage}
}}
\vspace{-0.17in}
\end{table*}

\begin{table}[]
\centering
\scalebox{0.7}{
\begin{tabular}{ccccccc}
    \Xhline{1pt}
    \multicolumn{3}{c}{Losses} &
    \multirow{2}{*}{Pose Error↓} & \multirow{2}{*}{Pose AUC↑} & \multirow{2}{*}{Mesh Error↓} & \multirow{2}{*}{Mesh AUC↑} \\
    \cline{1-3}
    ${E}_{UV}$ & ${E}_{grad}$ & ${E}_{verts}$\\
    \hline
    $\checkmark$ & & & 0.75 & 0.850 & 0.77 & 0.846\\
    $\checkmark$ & $\checkmark$ & & 0.73 & 0.854 & 0.75 & 0.850\\
    $\checkmark$ & $\checkmark$ & $\checkmark$ & \textbf{0.72} & \textbf{0.856} & \textbf{0.74} & \textbf{0.852}\\
    \Xhline{1pt}
\end{tabular}
}
\vspace{-0.1in}
\caption{Ablation studies for different losses used in our method on the FreiHAND testing set. 
}
\vspace{-0.2in}
\label{table:ablation}
\end{table}

\begin{table}[]
\scalebox{0.75}{
\begin{tabular}{c|c|cccc}
\Xhline{1pt}
\cellcolor[HTML]{FFFFFF}$f_{ac}$     & \cellcolor[HTML]{FFFFFF}Crop Ratio & \cellcolor[HTML]{FFFFFF}Pose Error↓ & \cellcolor[HTML]{FFFFFF}Pose AUC↑ & Mesh Error↓   & Mesh AUC↑      \\
\hline
\rowcolor[HTML]{EFEFEF} 
\cellcolor[HTML]{FFFFFF}                      & \cellcolor[HTML]{FFFFFF}1          & 0.85                                & 0.831                             & 0.87          & 0.827          \\
\cline{2-6}
\rowcolor[HTML]{FFFFFF} 
\cellcolor[HTML]{FFFFFF}                      & 3/4                                & 0.81                                & 0.839                             & 0.82          & 0.837          \\
\cline{2-6}
\rowcolor[HTML]{DAE8FC} 
\multirow{-3}{*}{\cellcolor[HTML]{FFFFFF}w/o} & \cellcolor[HTML]{FFFFFF}1/2        & 0.76                                & 0.848                             & 0.78          & 0.845          \\
\hline
\rowcolor[HTML]{EFEFEF} 
\cellcolor[HTML]{FFFFFF}                      & \cellcolor[HTML]{FFFFFF}1          & 0.77                                & 0.847                             & 0.79          & 0.843          \\
\cline{2-6}
\rowcolor[HTML]{FFFFFF} 
\cellcolor[HTML]{FFFFFF}                      & 3/4                                & \textbf{0.72}                       & \textbf{0.856}                    & \textbf{0.74} & \textbf{0.852} \\
\cline{2-6}
\rowcolor[HTML]{DAE8FC} 
\multirow{-3}{*}{\cellcolor[HTML]{FFFFFF}w/}  & \cellcolor[HTML]{FFFFFF}1/2        & 0.74                                & 0.852                             & 0.76          & 0.849\\
\Xhline{1pt}
\end{tabular}
}
\vspace{-0.1in}
\caption{Comparison of the affine connection module $f_{ac}$ used or not, where three crop ratios are compared in each case.}
\vspace{-0.2in}
\label{table:ac_ablation}
\end{table}

\begin{table*}[tb]
\centering
\makebox[0pt][c]{\parbox{1\textwidth}{
\begin{minipage}[h]{0.45\textwidth}
    \centering
    \begin{minipage}[t]{0.9\textwidth}
        \centering
        \setlength{\abovecaptionskip}{0.cm}
        \scalebox{0.8}{
        \begin{tabular}{ccc}
        \Xhline{1pt}
        Variants of our Method & AUC of PCK↑    & AUC of PCV↑    \\
        \hline
        UV1   & 0.860  & 0.856 \\
        UV2  & 0.862  & 0.860 \\
        UV3  & \textbf{0.865} & \textbf{0.862} \\
        \Xhline{1pt}
        \end{tabular}
        }
        \vspace{0.1cm}
        \caption{The results comparison of different UV forms in different stages.
        }
        \label{table:3}
    \end{minipage}%
    \vspace{0.2cm}
    \par
    \begin{minipage}[t]{0.9\textwidth}
        \centering
        \setlength{\abovecaptionskip}{0.cm}
        \scalebox{0.8}{
        \begin{tabular}{ccc}
        \Xhline{1pt}
        Variants of our Method & AUC of PCK↑    & AUC of PCV↑    \\
        \hline
        S0                 & \textbf{0.865} & \textbf{0.862} \\
        S1                 & 0.862          & 0.858\\
        S2                 & 0.839          & 0.834          \\
        \Xhline{1pt}
        \end{tabular}
        }
        \caption{The results comparison of different UV forms in different stages, $S_i$ means the i-th pyramid level output $I_{UV}^i$.
        }
        \label{table:reso}
    \end{minipage}%
\end{minipage}
\begin{minipage}[h]{0.55\textwidth}
\includegraphics[width=0.49\linewidth]{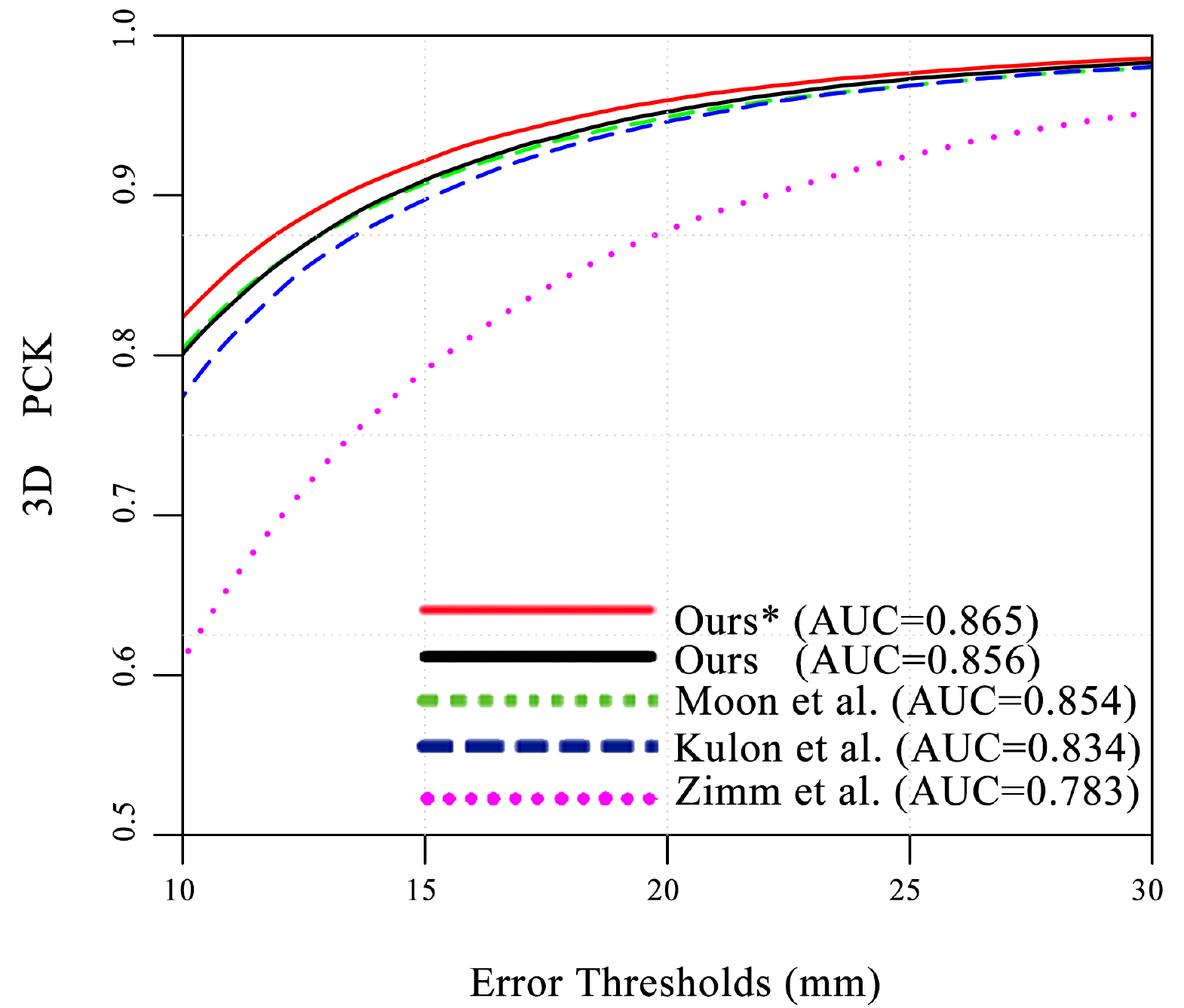}
\includegraphics[width=0.49\linewidth]{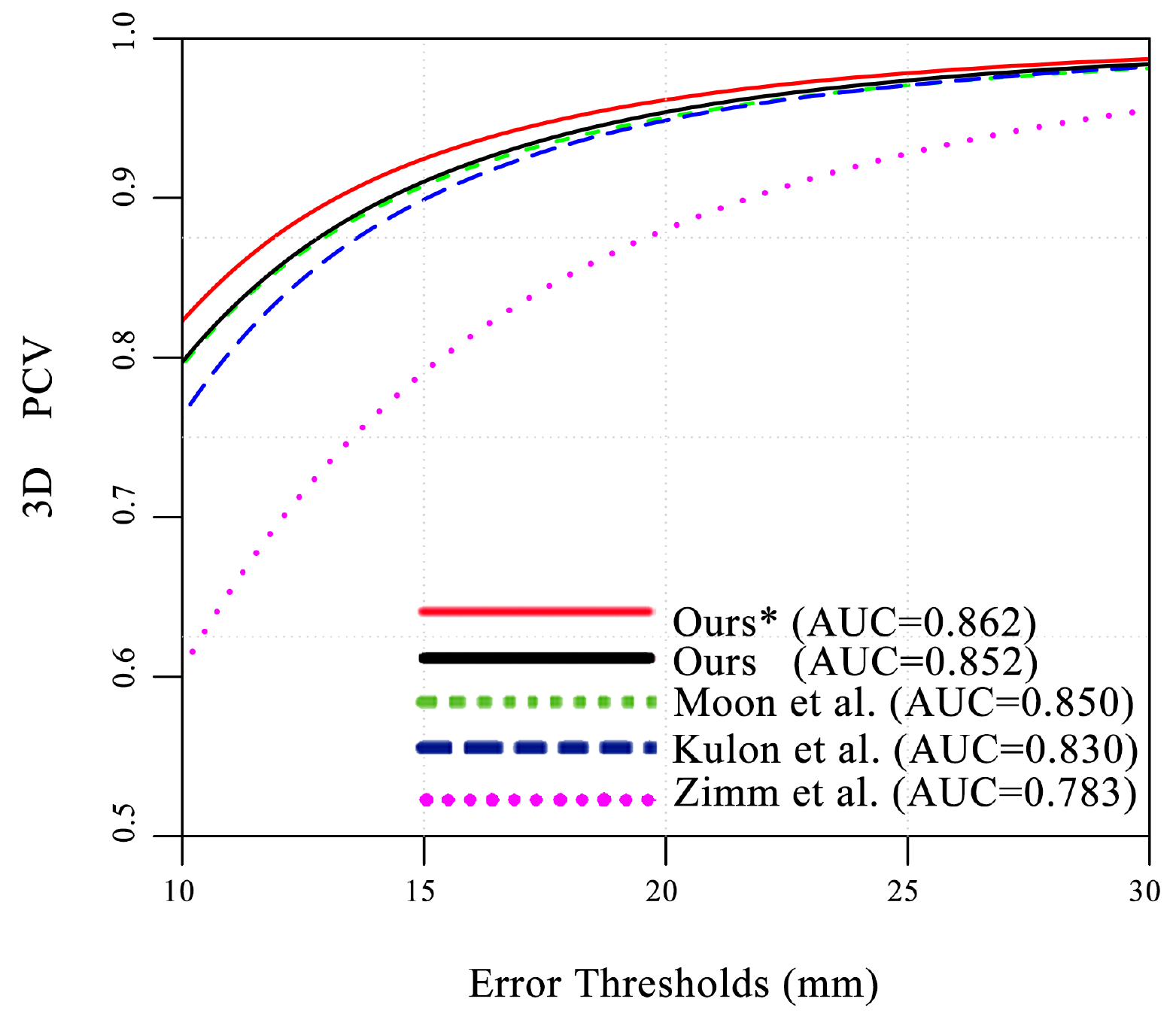}
\captionof{figure}{Comparison of 3D PCK and 3D PCV on the FreiHAND dataset. The proposed method is superior to \cite{Kulon_2020_CVPR},  \cite{moon2020i2l} and \cite{zimmermann2019freihand}.}
\label{fig:4}
\end{minipage}
}}
\vspace{-0.1in}
\end{table*}

\subsection{Implementation}
\label{sec:implementation}
We train our model on four NVIDIA Tesla V100 GPUs. Adam \cite{kingma2014adam} is used to optimize the network and PyTorch \cite{paszke2019pytorch} is used for implementation. The proposed model consists of two parts, i.e., the AffineNet and the SRNet, where the AffineNet aims to reconstruct hand mesh at MANO model level and the SRNet is designed to predict hand mesh with more detail. No image set provides both MANO-level and super MANO-level mesh annotations. Thus we adopt a stage-wise training strategy to optimize the network modules by using different data supervision for different parts, i.e., use the image dataset (such as FreiHAND) to train the AffineNet while use scan data (such as SHS) to train the SRNet.

The AffineNet is trained for 200 epochs with the batch size of 128 and the learning rate initialized to $1\times10^{-4}$ and changed according to a Cosine Learning rate decay. The input image is cropped to $3\times256\times256$. During its training, the input image is augmented by scaling, rotation and, color channel permutation. The SRNet is trained for 100 epochs with the batch size of 512 and the learning rate is set to $1\times10^{-3}$. The input and output UV position map of SRNet is $3\times256\times256$. 
For each sample in the SHS dataset, there are two sources of SRNet's input, one is the UV of the corresponding MANO mesh, and the other is the UV map after Gaussian smoothing.
\subsection{Comparison with State-of-the-art Methods}
\label{sub:sota}
Since our SRNet part is trained by scanning data, we only compare the results of AffineNet with other methods to ensure the fairness of the comparison.
We compare the proposed method with several state-of-the-art approaches \cite{boukhayma20193d,choi2020pose2mesh, hasson2020leveraging, Kulon_2020_CVPR,  moon2020i2l,zimmermann2019freihand} for hand pose and shape estimation on the \mbox{FreiHAND} dataset.
The results are shown in Table~\ref{table:1}. In general, methods without directly regressing MANO parameters (\cite{choi2020pose2mesh,Kulon_2020_CVPR,moon2020i2l} and our method) perform better than methods using MANO parameters regression (\cite{boukhayma20193d,hasson2019learning,zimmermann2019freihand}).
When no extra training data is used, our method surpasses all previous methods whether or not they use MANO model prior. When the additional training data (the training set of ObMan and YT-3D) is used, our method achieves better performance (see ``Ours* (AffineNet)" in the table). We further plot the 3D PCK and PCV of the FreiHAND test set compared with some state-of-the-art methods \cite{Kulon_2020_CVPR,moon2020i2l,zimmermann2019freihand}, where our method shows better performance.

In the hand-object interaction scenario, we compare with state-of-the-art methods \cite{hampali2020honnotate,hasson2020leveraging,hasson2019learning} on the HO3D dataset in Table~\ref{table:ho3d}. In the condition that no extra training data is used, our method outperforms all previous methods.
In addition, when the additional training data (the training set of FreiHAND, ObMan, and YT-3D) is used, our method achieves better performance (see ``Ours* (AffineNet)" in the table).
We also notice that our cross-dataset evaluation results (\mbox{``Ours$\dagger$ (AffineNet)"} in the table) surpass other results, where different from their training on the HO3D training set, we train the model on a combination of data from other datasets (the training data of FreiHAND, ObMan, and YT-3D).
Even though most of the samples in the training set (FreiHAND, ObMan, and YT-3D) represent bare hands while the samples in the HO3D test set are hand-object interaction images, this cross-dataset evaluation still has good performance, showing the robustness and effectiveness of the proposed AffineNet.



\subsection{Evaluation of Hand Super-resolution Reconstruction}
\label{sub:sr_evla}
The SRNet is trained on the SHS dataset, and the training detail is shown in Section~\ref{sec:implementation}. Once trained, our SRNet can be directly used for hand super-resolution reconstruction without any fine-tuning. Here, we evaluate the SRNet on FreiHAND, HO3D and HIC. 

The FreiHAND and HO3D are annotated using MANO fitting, thus their ground truth meshes are as coarse as MANO hand mesh. We use the output UV position maps of AffineNet as the input of the SRNet, and the SRNet can output hand meshes with higher fidelity. 
The qualitative results on HO3D and FreiHAND are shown in Figure~\ref{fig:pipeline}. For each input RGB image, we visualize the output of the AffineNet, the output of the SRNet in two viewpoints. We find that the SRNet outputs higher-resolution hand meshes than the output of the AffineNet while preserving the same pose information. The high-resolution meshes of the SRNet show
smoother and more realistic skin surface.


To quantitatively evaluate the SRNet, experiments are conducted on the HIC dataset. The input of the SRNet is the UV position map converted from its MANO-fitted mesh with 778 vertices (``Input" in Table~\ref{table:srnet}), and the output is the high-resolution UV position map that can be converted into a high-fidelity mesh with 3093 vertices (``Output" in the table). Besides, we also down-sample the high-fidelity mesh to 778 vertices (``Output (sampled)" in the table). 
In order to compare these three sets of hand surfaces, we render them as depth maps based on the viewpoint of the input image and use the depth map observation to calculate the RMSE and PSNR of the rendered depth image. Note that the background is erased by the intersection of these depth maps.
As shown in Table~\ref{table:srnet}, the output of SRNet (``Output" in the table) shows a better reconstruction quality to the original depth map. We also notice that the down-sampled output (``Output (sampled)" in the table) obtains higher accuracy than the mesh fitted by MANO (``Input" in the table), even though they have the same mesh resolution.

\subsection{Ablation Study}
\label{sub_sec:ablation}
\noindent \textbf{Effect of Each Loss Function.}\quad As presented in Table~\ref{table:ablation}, we give evaluation results of AffineNet on the FreiHAND test set of settings with different losses used during the training. From the table, we can see that the best performance is achieved when all proposed loss functions are used. More results are shown in the Appendix.\\
\noindent \textbf{Effect of the Affine Connection.}\quad As presented in Table~\ref{table:ac_ablation}, we give evaluation results on FreiHAND of settings with the affine connection module $f_{ac}$ used or not, where three crop ratios are compared in each case. Here, the cropping operation is designed to compare the effect of different foreground and background ratios on the reconstruction result. For example, when the crop ratio is 1/2, the image is center cropped by 1/2 of the width and height and then resized into the original size. In Table~\ref{table:ac_ablation}, for each crop ratio, we find the ``w/ $f_{ac}$" gets better performance than ``w/o $f_{ac}$" and it gets the best performance when the crop ratio is 3/4. Therefore, we choose 3/4 as the crop ratio in other experiments. In this case, $f_{ac}$ results in an 11.1\% reduction in pose error and a 9.8\% reduction in mesh error.

\noindent \textbf{Effect of UV unfolding forms.}\quad The UV map is obtained by unfolding the hand mesh. Thus, the valid part of the UV map is dependent on the crop trajectory of the surface (i.e., the black lines on the 3D surface in Figure~\ref{fig:2}). We compare three different UV position maps using different crop and combine schemes, and indicate UV1, UV2 and, UV2 from left to right in Figure~\ref{fig:2}.
UV1 separates the front and back of the hand and the area on the UV position map of each piece is proportional to the area of the mesh surface. UV2 and UV3 don’t separate each finger and the area on the UV position map of each piece is proportional to the vertex number of the mesh surface, where each piece has a different position in UV space.
In Table~\ref{table:3}, we give evaluation results on the FreiHAND test set while using a combined training set (refer to the train data of FreiHAND, ObMan, and YT-3D).
Despite different UV position maps are used, the performance of hand pose and shape estimation is similar, where less than 0.7\% difference on AUC of PCK/PCV is shown.
The results show that our method is robust to the UV position map designing. In this paper, we use UV3 as the template UV position map.

\noindent \textbf{Comparison of results from UV maps of different stages.}\quad As illustrated in Section~\ref{sec:loss_affine}, the supervision is used on multiple-scale UV maps. Thus, the mesh can be recovered from each UV map. Here, we compare the mesh prediction results of the last three UV maps (S0, S1, and S2) on the FreiHAND test set while using a combined training set. S0 indicates the full resolution UV map prediction, while S1 refers to the second last output with 1/2 of the full resolution and S2 refers to the third last output with 1/4 of the full resolution. As shown in Table~\ref{table:reso}, from low-resolution S2 to high-resolution S0, the AUC of PCK and PCV show obvious improvement. 
\section{Conclusion}
We have presented a novel I2UV-HandNet approach for accurate and high-fidelity 3D hand reconstruction from a single color image.
The proposed UV position map enables representing a 3D hand surface in an image style.
For accurate hand pose and shape estimation from the monocular images, we present an AffineNet to predict the UV position map from the RGB input. In AffineNet, a hierarchical coarse-to-fine regressing architecture is designed with a novel affine connection module that can resolve the coordinates ambiguity between the RGB image and the UV map.
The proposed AffineNet achieves state-of-the-art performances on multiple challenging datasets. 
For high-fidelity hand shape reconstruction, we present an SRNet to restore a high-resolution UV position map from a low-resolution one. The proposed SRNet is not likely to be affected by the reconstruction method, and can robustly restore a high-fidelity hand from the inputted coarse shape.
As for the future study, the UV-based hand representation can be extended to more complex joint hand-hand or hand-object reconstruction tasks, or the architecture can be modified for enabling sparse/weak supervision.


{\small
\normalem
\bibliographystyle{ieee_fullname}
\bibliography{arxiv}
}

\appendix
\noindent \textbf{\Large{Appendix}}
\\

In this appendix, we detail our network architecture in Section~\ref{sec:imple_archit}; in Section~\ref{sec:quali}, we additionally provide more results.

\section{Network Architecture Details}
\label{sec:imple_archit}
We detail the network architecture in detail in Tables~\ref{table:encoder_size}-\ref{table:sr_archit}.
Table~\ref{table:encoder_size} describes the size of each feature map from the encoder. Here, we use five feature maps of ResNet-50 with a convolution operation using $3\times3$ convolution kernels and the output channels are 64, 128, 256, 512, and 1024. 
The feature maps are then fed to the decoder of the proposed AffineNet which predicts UV position maps in multiple resolutions, as detailed in Table~\ref{table:de_archit}.
We note that ``affine-operation" indicates an affine transformation from $E^i$ to $A^i$ according to the current UV map prediction $I^i_{UV}$. Along the expansive path, the UV position map gets more and more accurate (from the coarsest $I^4_{UV}$ to the most accurate $I^0_{UV}$), thus the $A^i$ gets better accurate spatial alignment, then resulting in more accurate predicted UV position map of the next level $I^{i-1}_{UV}$. 
In conclusion, the prediction of the UV position map and the degradation of coordinate ambiguities are coarse-to-fine and mutually improved.
The SRNet is described in Table~\ref{table:sr_archit}. 

\begin{table}[h]
\centering
\begin{tabular}{|c|p{3.5cm}<{\centering}|}
\Xhline{1pt}
\rowcolor[HTML]{DCDCDC} 
Encoder Feature Map & Size         \\ \Xhline{1pt}
Input               & (3,256,256)  \\ \hline
$E^1$                  & (64,128,128) \\ \hline
$E^2$                  & (128,64,64)  \\ \hline
$E^3$                  & (256,32,32)  \\ \hline
$E^4$                  & (512,16,16)  \\ \hline
$E^5$                  & (1024,8,8)   \\ \Xhline{1pt}
\end{tabular}
\caption{The size of each feature map in the encoder.}
\vspace{-0.3in}
\label{table:encoder_size}
\end{table}

\begin{table*}[]
\centering
\begin{tabular}{|p{3cm}<{\centering}|p{5cm}<{\centering}|p{2.5cm}<{\centering}|p{3.5cm}<{\centering}|}
\Xhline{1pt}
\rowcolor[HTML]{DCDCDC} 
Input                          & Operation                                     & Output & Output Size  \\
\Xhline{1pt}
$E^5$  & Upsample/Conv/BN/ReLU & $D^4$   & (512,16,16)  \\
\hline
\rowcolor[HTML]{CEFBFA} 
$D^4$ & Conv/Sigmoid & $I_{UV}^4$ & (3,16,16)\\
\hline

$E^4$,$I_{UV}^4$ & Affine-operation/Upsample & $A^3$ & (512,32,32)\\
\hline
$D^4$ & Upsample/Conv/BN/ReLU & $D^3$ & (256,32,32)\\
\hline
$I_{UV}^4$ & Upsample & $\hat{I}_{UV}^3$ & (3,32,32)\\
\hline
$A^3$,$D^3$,$\hat{I}_{UV}^3$ & Concat/Conv/BN/ReLU & $D'^3$ & (256,32,32)\\
\hline
\rowcolor[HTML]{CEFBFA} 
$D'^3$ & Conv/Sigmoid & $I_{UV}^3$ & (3,32,32)\\
\hline

$E^3$,$I_{UV}^3$ & Affine-operation/Upsample & $A^2$ & (256,64,64)\\
\hline
$D'^3$ & Upsample/Conv/BN/ReLU & $D^2$ & (128,64,64)\\
\hline
$I_{UV}^3$ & Upsample & $\hat{I}_{UV}^2$ & (3,64,64)\\
\hline
$A^2$,$D^2$,$\hat{I}_{UV}^2$ & Concat/Conv/BN/ReLU & $D'^2$ & (128,64,64)\\
\hline
\rowcolor[HTML]{CEFBFA} 
$D'^2$ & Conv/Sigmoid & $I_{UV}^2$ & (3,64,64)\\
\hline

$E^2$,$I_{UV}^2$ & Affine-operation/Upsample & $A^1$ & (128,128,128)\\
\hline
$D'^2$ & Upsample/Conv/BN/ReLU & $D^1$ & (64,128,128)\\
\hline
$I_{UV}^2$ & Upsample & $\hat{I}_{UV}^1$ & (3,128,128)\\
\hline
$A^1$,$D^1$,$\hat{I}_{UV}^1$ & Concat/Conv/BN/ReLU & $D'^1$ & (64,128,128)\\
\hline
\rowcolor[HTML]{CEFBFA} 
$D'^1$ & Conv/Sigmoid & $I_{UV}^1$ & (3,128,128)\\
\hline

$E^1$,$I_{UV}^1$ & Affine-operation/Upsample & $A^0$ & (64,256,256)\\
\hline
$D'^1$ & Upsample/Conv/BN/ReLU & $D^0$ & (32,256,256)\\
\hline
$I_{UV}^1$ & Upsample & $\hat{I}_{UV}^0$ & (3,256,256)\\
\hline
$A^0$,$D^0$,$\hat{I}_{UV}^0$ & Concat/Conv/BN/ReLU & $D'^0$ & (32,256,256)\\
\hline
\rowcolor[HTML]{CEFBFA} 
$D'^0$ & Conv/Sigmoid & $I_{UV}^0$ & (3,256,256)\\
\hline

\Xhline{1pt}
\end{tabular}
\caption{Layer specification for the decoder part. ``Upsample" indicates to use bilinear interpolation to enlarge the spatial size by 2 times; ``Conv" indicates convolution operation using $3\times3$ convolution kernel with zero padding; ``BN" indicates batch normalization.}
\label{table:de_archit}
\end{table*}

\begin{table*}[]
\centering
\begin{tabular}{p{2cm}<{\centering}|p{2cm}<{\centering}|p{2.5cm}<{\centering}|p{2.5cm}<{\centering}|p{2cm}<{\centering}|p{2cm}<{\centering}}
\rowcolor[HTML]{DCDCDC} 
\Xhline{1pt}
Layer   & Operation & Input Size   & Output Size  & Kernel Size & Padding \\ \Xhline{1pt}
\rowcolor[HTML]{FFFFFF} 
Level 1 & Conv/ReLU & (3,256,256)  & (64,256,256) & (9,9)       & (4,4)   \\ \hline
\rowcolor[HTML]{FFFFFF} 
Level 2 & Conv/ReLU & (64,256,256) & (32,256,256) & (5,5)       & (2,2)   \\ \hline
\rowcolor[HTML]{FFFFFF} 
Output  & Conv      & (32,256,256) & (3,256,256)  & (5,5)       & (2,2)  \\ \hline
\Xhline{1pt}
\end{tabular}
\caption{Layer specification for the SRNet.}
\label{table:sr_archit}
\end{table*}

\section{Additional Qualitative Results}
\label{sec:quali}
\begin{figure*}
    \centering
    \includegraphics[width=1\linewidth]{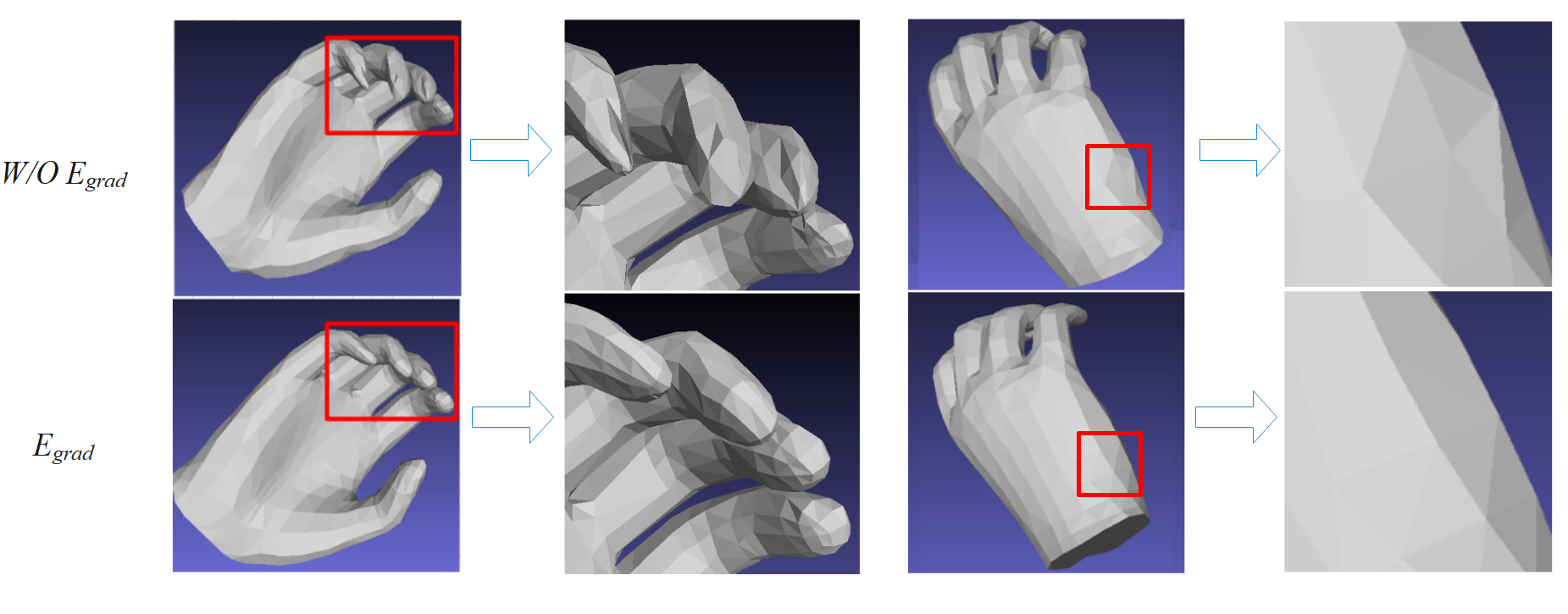}
    \caption{Qualitative comparison of 3D hand reconstruction results from AffineNet in terms of $E_{grad}$ used or not.}
    \label{fig:e_grad}
\end{figure*}

\begin{figure*}
    \centering
    \includegraphics[width=1\linewidth]{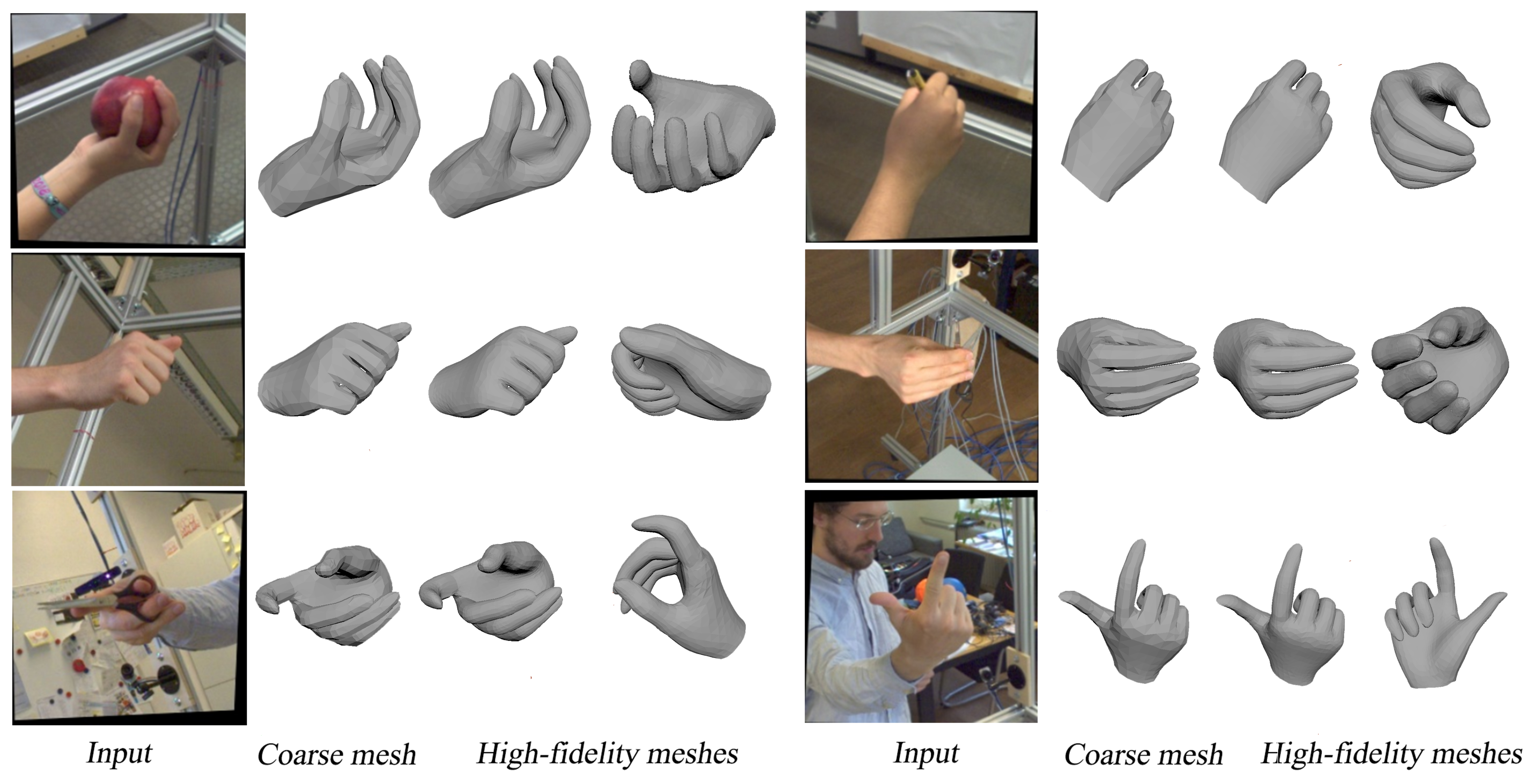}
    \caption{Qualitative visualization of our method on the FreiHAND testing set.}
    \label{fig:supp_fh}
\end{figure*}

\begin{figure*}
    \centering
    \includegraphics[width=1\linewidth]{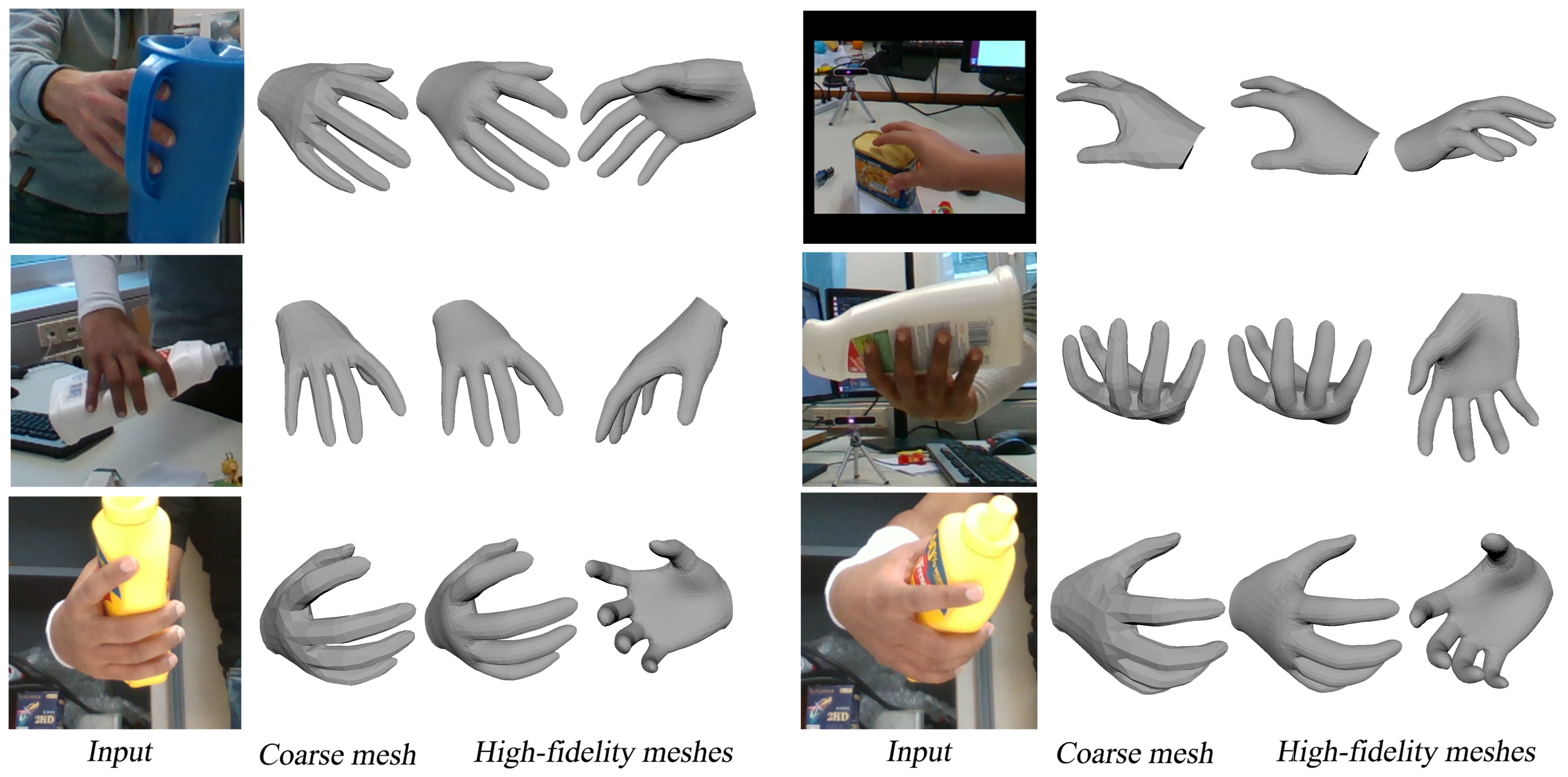}
    \caption{Qualitative visualization of our method on the HO3D testing set.}
    \label{fig:supp_ho3d}
\end{figure*}

\begin{figure*}[t]
    \centering
    \includegraphics[width=1\linewidth]{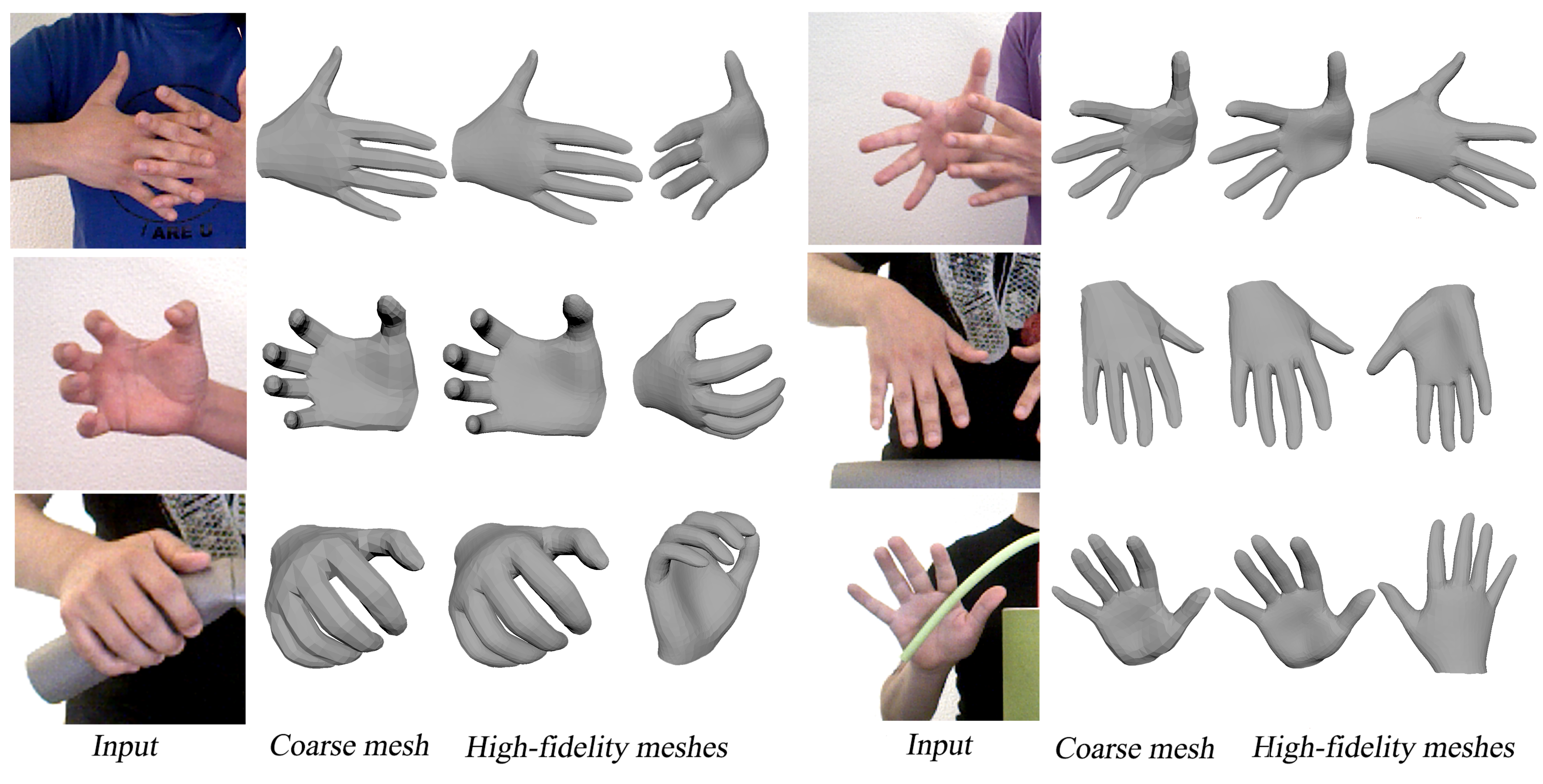}
    \caption{Qualitative visualization of our method on the HIC dataset.}
    \label{fig:supp_hic}
    \vspace{4in}
\end{figure*}

We show additional evaluate on the impact of $E_grad$ in Figure~\ref{fig:e_grad} (also see Section~4.6 of the main text).
As shown in the figure, $E_ {grad}$ makes the mesh have better surface continuity, that is, smoother. The two samples that do not use $E_ {grad}$ have unreasonable folds or depressions on their surfaces.
We show additional visualization of our predictions via AffineNet and SRNet for FreiHAND test set \cite{zimmermann2019freihand} (see Figure~\ref{fig:supp_fh}), HO3D test set \cite{hampali2020honnotate} (see Figure~\ref{fig:supp_ho3d}), and HIC dataset \cite{tzionas2016capturing} (see Figure~\ref{fig:supp_hic}).
Here, the AffineNet is trained on a combination of data (the training data of FreiHAND, Obman, and YT-3D) and the SRNet is trained on the SHS dataset. The visualization shows our method outputs accurate and high-fidelity hand reconstruction results, and the results on HO3D and HIC show that our network has good generalization.

\end{document}